\documentclass[letterpaper, 10 pt, conference]{ieeeconf}  % Comment this line out if you need a4paper

\usepackage{times}
\usepackage[numbers]{natbib}
\usepackage{multicol}
\usepackage[bookmarks=true]{hyperref}

\usepackage{lipsum}
\usepackage{comment}
\usepackage{balance}
\usepackage{graphicx}
\usepackage[table]{xcolor}
\usepackage{caption}
\usepackage{listings}

\usepackage{seqsplit}

\IEEEoverridecommandlockouts                              % This command is only needed if 
\usepackage{tabularx} 
\title{\Large \bf
Sequential Discrete Action Selection via Blocking Conditions and Resolutions
}

\author{Liam Merz Hoffmeister$^{1}$, Brian Scassellati$^{1}$, Daniel Rakita$^{1}$
\thanks{$^{1}$Liam Merz Hoffmeister, Brian Scassellati, and Daniel Rakita are with the Department of Computer Science, Yale University,
        New Haven, CT 06520, USA
        {\tt\small liam.merzhoffmeister@yale.edu}}%
\thanks{This work was supported by Office of Naval Research award N00014-24-1-2124}
}

\begin{document}

\maketitle
\thispagestyle{empty}
\pagestyle{empty}

%%%%%%%%%%%%%%%%%%%%%%%%%%%%%%%%%%%%%%%%%%%%%%%%%%%%%%%%%%%%%%%%%%%%%%%%

\begin{abstract}
In this work, we introduce a strategy that frames the sequential action selection problem for robots in terms of resolving \textit{blocking conditions}, i.e., situations that impede progress on an action en route to a goal.  This strategy allows a robot to make one-at-a-time decisions that take in pertinent contextual information and swiftly adapt and react to current situations.  We present a first instantiation of this strategy that combines a state-transition graph and a zero-shot Large Language Model (LLM).  The state-transition graph tracks which previously attempted actions are currently blocked and which candidate actions may resolve existing blocking conditions.  This information from the state-transition graph is used to automatically generate a prompt for the LLM, which then uses the given context and set of possible actions to select a single action to try next.  This selection process is iterative, with each chosen and executed action further refining the state-transition graph, continuing until the agent either fulfills the goal or encounters a termination condition.  We demonstrate the effectiveness of our approach by comparing it to various LLM and traditional task-planning methods in a testbed of simulation experiments.  We discuss the implications of our work based on our results.
\end{abstract}

\section{Introduction}
\label{sec:introduction}

Robots offer the promising potential to assist people across a variety of domains, including home care, healthcare, agriculture, and manufacturing.  In these scenarios, robots are tasked with determining the most beneficial action to take at any given moment to achieve specific goals or objectives.  

Much previous work has investigated the action selection problem posed above \cite{fikes1971strips, helmert2006fast, hoffmann2005contingent, hoffmann2001ff, mcdermott1998pddl}.  At a high level, previously proposed algorithms often reason over the broad space of which actions are \textit{viable} at a certain state.  For instance, task planners often consider \textit{prerequisites} that define whether a given action is viable, then use modeled effects to define how the environment would change if a certain action is executed.  Using this information, a task planner considers many possible roll-outs of legal action trajectories, stopping when a sequence of actions reaches a given goal.          

The action viability-centered view of task selection has afforded several useful algorithms (\S\ref{sec:related_works}); however, the strategy as a whole can often lead to computational and representational challenges.  For instance, if there are many viable actions that beget many viable actions to consider at a next step, and so on, the number of possible action branches scale exponentially with respect to the number of planning steps, often overwhelming computational resources.  While heuristics can help alleviate this issue \cite{helmert2006fast, hoffmann2001ff}, it may still be difficult to prune large segments of or find shortcuts through these dense branches of actions as it is often unclear which current actions will be necessary to achieve future goals.  Also, given the uncertainty of the real-world, it can often be impossible to perfectly model which actions will be viable at what time.  For instance, it is impossible to know if the robot will know where a target object is or if it will have the physical strength to lift a certain object ten steps into the future.  Thus, if any modeled assumption is broken in the real world, a full re-plan often has to be triggered, exacerbating the computational challenges outlined above.                   

In this work, we introduce a strategy that frames the action selection problem for robots in terms of resolving \textit{blocking conditions}, i.e., situations that impede progress on an action en route to a goal.  This strategy allows a robot to make one-at-a-time decisions that take in pertinent contextual information and swiftly adapt and react to current situations.  

Our proposed strategy follows three high-level steps: (1) The robot \textit{selects an action} from a set of candidate actions; (2) The robot assesses if any conditions are \textit{blocking progress} on the current action; and (3) If progress is blocked on its current action, the robot evaluates all attempts to \textit{resolve} the situation, incorporating possible resolutions into its set of candidate actions (circling back to Step 1). At this juncture, the robot may either choose an action that resolves the issue and resume the previously blocked action or prioritize a new action, reflecting an adapted strategy.  This loop continues until the agent fulfills the goal criteria or encounters a termination condition. 

\begin{figure}[t!]
	\includegraphics[width=\columnwidth]{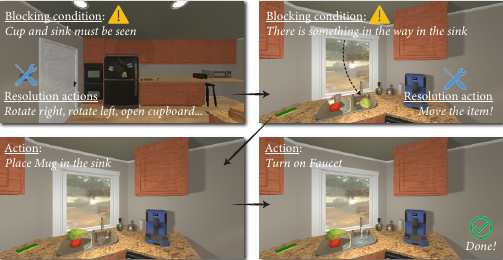}
	\caption{ We present a strategy that frames action selection in terms of resolving \textit{blocking conditions}, i.e., situations that impede progress on an action en route to a goal.  This example illustrates a robot using blocking conditions and resolutions to complete the task ``wash the cup in the sink'' in the AI2Thor simulation environment.  This task is part of our evaluation. }
	\label{fig:teaser}
	\vspace{-15pt}
\end{figure}

Our central premise is that centering action selection decisions around blocking conditions and resolutions will ease the computational and representational challenges associated with discrete action selection, ultimately enabling more successful robot task execution.  For instance, the exponential action branching challenge discussed above is eased by only considering actions that can either directly achieve the given goal or resolve a blocking condition currently impeding progress toward the goal.  Our empirical results (\S\ref{sec:evaluation}) suggest that this set of candidate actions will be much smaller than the set of all viable actions at a particular state.  Also, because action decisions are made one at a time in response to blocking conditions, it is straightforward to quickly adapt strategies when needed in the face of physical obstructions or lack of information that could not be modeled \textit{a priori}.    

We present a first instantiation of our strategy that combines a state-transition graph and a zero-shot Large Language Model (LLM).  The state-transition graph tracks which previously attempted actions are currently blocked and which candidate actions may resolve existing blocking conditions (Steps 2 and 3 in the strategy above).  This information from the state-transition graph is used to automatically generate a prompt for the LLM, which then uses the given context and set of candidate actions to select a single action to try next (Step 1 in the strategy above).  This selection process is iterative, with each chosen and executed action further refining the state-transition graph.  The algorithm continues iterating until the agent fulfills the goal criteria or encounters a termination condition. 

We demonstrate the effectiveness of our approach by comparing it to various LLM and traditional task-planning methods in a testbed of simulation experiments (\S\ref{sec:evaluation}).  We conclude by discussing the implications and limitations of our work based on our results.  We provide open-source code for an implementation of our approach.\footnote{\href{https://github.com/Apollo-Lab-Yale/llm_task_planning}{https://github.com/Apollo-Lab-Yale/llm\_task\_planning}}

\section{Related Works}
\label{sec:related_works}

\subsection{Task Planning}

The goal of task planning is to compute a sequence of actions that achieve some given goal.  Task planners typically reason over some logic-based domain language, such as STRIPS \cite{fikes1971strips} or PDDL \cite{mcdermott1998pddl}.  At a high level, these domain languages specify a start state, a goal state, and what constitutes legal, viable actions at any given state.  

Over the years, several highly efficient task planning algorithms have been developed to find feasible sequences of actions from start to goal, including Fast Forward (FF) \cite{hoffmann2001ff} and Fast Downward (FD) \cite{helmert2006fast}.  FF uses a graph search strategy coupled with a carefully designed heuristic that estimates the cost to reach the goal from the current state.  FD decomposes planning tasks into more manageable sub-problems, allowing for significant speedups on a hierarchical representation.  

% Both of these approaches have significantly advanced the domain of automated planning; traditional planners of this class necessitate a well-defined, observable domain for optimal functionality.   

% More recently, symbolic planners have been combined with sampling-based planners 

% Task planning has seen significant progress, starting with the pioneering efforts of \citet{fikes1971strips}, who introduced a symbolic representation of the environment in \textit{Strips: A new approach to the application of theorem proving to problem-solving}. This framework laid the groundwork for developing PDDL (The Planning Domain Definition Language) \cite{mcdermott1998pddl}, which standardized how planning problems and domains are defined.

% Innovations continued with the integration of symbolic planning and sampling methods, as showcased by Garrett et al. in \textit{Pddlstream: Integrating symbolic planners and black box samplers via optimistic adaptive planning} \cite{garrett2020pddlstream}, facilitating planning within fully observable domains. Efficiencies in symbolic planning were notably advanced through the introduction of systems like the FF: The Fast-Forward Planning System \cite{hoffmann2001ff} and the Fast Downward Planning System \cite{helmert2006fast}. However, these systems still require precise, observable domains.

Several task planners were subsequently presented that address the challenges of non-observability and uncertainties.  For instance, FF-Replan \cite{yoon2007ff}, which modified the FF planner to address non-deterministic outcomes through a ``determinize-and-replan'' strategy, marked a significant shift towards managing unpredictable elements in task planning, although it did not completely resolve the issue of state uncertainties.

Further advancement was achieved through planning over belief states, i.e., some representation of many possible world states \cite{bertoli2002improving, bonet2000planning}.  For instance, \citet{hoffmann2005contingent} presents implicit belief state planning where actions can have non-deterministic outcomes.  Handling implicit belief states allows the planner to consider various contingencies and develop plans that are robust to uncertainties.

% Nevertheless, the challenge of fully representing such uncertainties within planning domains remained, highlighting the complexity of effectively integrating uncertainty into deterministic planning frameworks.

% Task planning began to directly address non-observability and uncertainties with Yoon et al.'s FF-Replan \cite{yoon2007ff}, adapting the FF system to manage non-deterministic outcomes by employing a ``determinize-and-replan strategy''. However, this approach did not fully accommodate state uncertainties.  Hoffmann and Brafman's work on Contingent Planning via Heuristic Forward Search with Implicit Belief States \cite{hoffmann2005contingent} marked a significant step in accommodating initial state uncertainty and action non-determinism, albeit with the challenge of representing such uncertainty in planning domains fully.

Our proposed strategy draws on many concepts from task planning.  For instance, our strategy uses a logic-based domain language to specify goals and track information about the world and possible actions.  However, our work differs in that instead of framing search in terms of what actions are viable from a certain state (or belief state), our strategy frames action selection in terms of which actions may resolve blocking conditions that have been observed.  This strategy enables a robot to make one-at-a-time decisions that take in contextual information and adapt and react to current situations.  

\subsection{LLMs for Planning}

% Recent explorations into the convergence of Large Language Models (LLMs) and planning have illuminated LLMs' potential to enhance Automated Planning and Scheduling (APS), especially through their ability to interpret natural language instructions and generate plans aligned with the given context \cite{LLMsAPSProspects}.

Recent developments have demonstrated remarkable progress in natural language processing models, particularly Large Language Models (LLMs) \cite{zhao2023survey}.  One key area of exploration is the potential for these models to understand a planning query in natural language and generate a clear, step-by-step response to achieve a specific goal.  

% This area of work is compelling as communication via natural language is intuitive and accessible, and producing responses as a sequence of string tokens scales linearly with the length of the output.  

% This scaling is in stark contrast to the exponential scaling often seen in task planning associated with the branching factor of potential actions originating from any given state.

Modern LLMs are known to struggle with various planning-related tasks \cite{bubeck2023sparks}.  However, recent work has investigated this problem.  For instance, work from \citet{ahn2022i} and \citet{chalvatzaki2023learning} have demonstrated how LLMs can bridge the gap between abstract instructions and actionable tasks, leveraging their vast semantic understanding. These works resonate with our emphasis on adaptability and dynamic action selection in response to evolving environments.  Also, the work by \citet{song2023llmplanner} and \citet{singh2022progprompt} emphasizes the role of LLMs in generating dynamic plans, grounding them in the physical environment, and employing programmatic prompting to aid in plan formation.

% LLM-Planner: Few-Shot Grounded Planning for Embodied Agents with Large Language Models \cite{song2023llmplanner} and ProgPrompt , which 

Frameworks combining LLMs with structured planning, such as the work by \citet{liu2023llmp} and \citet{ding2022robot}, showcase the benefits of integrating LLM-derived insights to navigate unforeseen scenarios and improve flexibility in changing settings.  Our first instantiation of our proposed strategy (outlined in \S\ref{sec:technical_details}) fits well in this category.  Our idea involves merging the contextual understanding provided by LLMs with the clarity and precision of a state-action graph representation.  This integration aims to create a flexible and robust method for action selection, combining the best of both to enhance decision-making processes. 

% Further contributions include LLM-Planner: Few-Shot Grounded Planning for Embodied Agents with Large Language Models \cite{song2023llmplanner} and ProgPrompt \cite{singh2022progprompt}, which emphasize the role of LLMs in generating dynamic plans, grounding them in the physical environment, and employing programmatic prompting to aid in plan formation.

% \subsection{MIP based Controllers}

% The discussions around Model Predictive Control (MPC) and Adaptive Model Predictive Control (AMPC) underscore their versatility and the computational hurdles encountered in real-time deployments \cite{qin1997overview, 9919375}. These discussions highlight the ongoing search for adaptive control strategies that can efficiently navigate real-time challenges and remain resilient against environmental changes, aligning with our approach's focus on adaptability and responsiveness in task planning and execution.
\section{Strategy Overview}
\label{sec:strategy_overivew}

In this section, we describe the high level strategy proposed in this work: framing discrete action selection around blocking conditions and resolutions.  In \S\ref{sec:technical_details}, we describe in detail one instantiation of this strategy which we subsequently use in our evaluation.   
% We describe the components of our strategy, enumerate the steps that our strategy follows, then provide a short example.   

% Details regarding our first instantiation of this strategy are found in \S\ref{sec:technical_details}.  

\begin{figure*}[t]
    % \vspace{-5pt}
    \includegraphics[width=\textwidth]{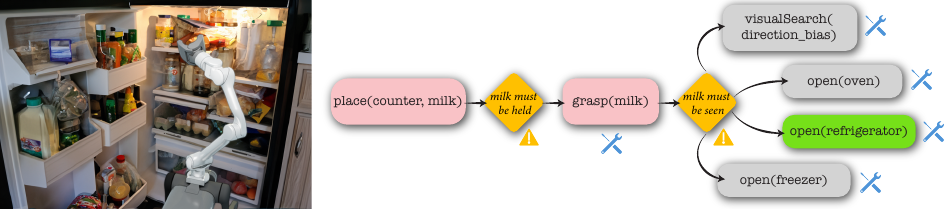}
    % \vspace{-20pt}
    \caption{\small The tree structure during the retrieving milk from the refrigerator example.  The red nodes signify blocked actions, the orange diamonds signify blocking conditions, gray nodes signify available actions, the green node is the currently selected action, and the blue symbols signify resolution actions that may fix the parent blocking condition. }
    \vspace{-15pt}
    \label{fig:tree_structure}
\end{figure*}

\subsection{Strategy Components}
\label{sec:strategy_components}

Our strategy builds on the following components.  Several of these components are also used in standard task planners, and are named the same, accordingly.

\begin{itemize}
    \item A set of distinct nouns in the world called \textit{instances}.
    \item A set of binary-valued functions of one or more instances, called \textit{predicates}. Each application of a predicate to a specific set of instances can return true (called a \textit{positive literal}) or false (called a \textit{negative literal}).  For example, suppose we have a predicate \texttt{Under} that takes two instances as arguments \texttt{\allowbreak{}Under(dirt, rug)}. 
   This predicate would return true and be a positive literal if dirt is under the rug.
    \item The current setting of all literals over all instances will be called the \textit{world state}.
    \item A set of \textit{actions} which each provide a functional specification (through predicates) for how the robot can change the world state.
    \item Each action has a set of \textit{effects}, which each specify the changes to the world state resulting from the execution of its action.  Importantly, our strategy allows for effects with truth values of \texttt{possibly\_true} or \texttt{possibly\_false} to accommodate uncertain results. 
    \item Each action has a set of \textit{blocking conditions}, which each provide reasons for why the action may not succeed.  For instance, example blocking modes for a ``pick up item'' action may be ``target object not in view'', ``item is out of reach'', or ``item is too heavy to lift''.
    \item Each blocking condition has a set of \textit{resolutions}, which are literals that must be satisfied for the blocking mode to be considered resolved.
    \item Each blocking condition has a corresponding set of \textit{resolution actions}, which are actions that have effects that would adjust the literals of the world to satisfy its resolutions.  
\end{itemize}

\subsection{Strategy Steps}
\label{sec:strategy_steps}

The strategy starts by representing some goals as a set of literals.  Actions with effects that directly achieve some aspect of these goals are added as roots of their own tree structures.  Then, our task execution via blocking conditions and resolutions strategy proceeds as follows:

(1) The robot \textit{selects an action} from a set of candidate actions.  The set of candidate actions is all leaf nodes in the tree structures mentioned above.  The module that makes this decision is called the \textit{selection engine}; (2) The robot attempts executing the selected action and assesses if any \textit{blocking conditions} are present on the current action; and (3) If the action executes successfully, it and all of its sibling nodes (i.e., nodes with the same parent) are removed from the tree structure.  If progress is blocked on the current action, the robot evaluates all attempts to \textit{resolve} the situation, adding all resolution actions for the blocking condition at hand as children nodes of the blocked action in the tree structure (circling back to Step 1).

At this juncture, the selection engine may either choose an action that resolves the issue in service of resuming the previously blocked action or prioritize a new action, reflecting an adapted strategy.  These three steps iterate until the robot either satisfies the goal literals or encounters a termination condition.    

\subsection{Strategy Example}
\label{sec:strategy_example}

Here, we outline how our strategy may apply to a robot tasked with retrieving milk from a refrigerator.  The goal literal for this task is \texttt{\allowbreak{}On(milk, counter) = true}.  

There is one action that has an effect that achieves the given goal: \texttt{\allowbreak{}Place(milk, counter)}.  The approach adds this action as a root node in a tree structure.  The selection engine chooses this action (it is the only option) and attempts to execute it.  However, a blocking condition is encountered: for an object to be placed, it must first be currently held.  The resolution literal for this blocking condition is \texttt{\allowbreak{}isHolding(milk) = true}.  

The robot has one possible way to elicit the desired effect: \texttt{\allowbreak{}grasp(milk)}.  This action is added as a child node to the \texttt{place} action node.  The action selection engine chooses the \texttt{\allowbreak{}grasp(milk)} action (again, the only option).  The robot attempts to execute this action, but another blocking condition is met: for an object to be grasped, it must currently be visible.  The resolution literal for this blocking condition is \texttt{\allowbreak{}isVisible(milk) = true}.

While no actions can guarantee the milk becoming visible, several actions \textit{may} lead toward achieving the desired outcome. For example, the actions \texttt{\allowbreak{}visualSearch(direction\_bias)}, \texttt{\allowbreak{open(\allowbreak{}refrigerator)}}, \texttt{\allowbreak{open(\allowbreak{}freezer)}}, \texttt{\allowbreak{open(\allowbreak{}oven)}} all have effects that include \texttt{\allowbreak{}isVisible(\allowbreak{}milk)} = \texttt{possibly\_true}.  

% These are all added to the tree structure, and \texttt{\allowbreak{}visual\allowbreak{}Search(direction\_bias)} is selected.  The robot executes the \texttt{\allowbreak{}visualSearch} action.  However, this strategy does not result in the milk being found.  This node is \textit{not} removed from the tree, however, as it is an action with a continuous parameter, \texttt{\allowbreak{}direction\_bias}, thus it is  impossible to exhaust its parameter space.

The selection engine chooses \texttt{\allowbreak{}open(refrigerator)}.  The tree structure at this point in the task is visualized in Figure \ref{fig:tree_structure}.  Upon successfully opening the refrigerator, new instances become visible inside, such as yogurt, juice, etc. However, the milk is not immediately detected. 

% The system then evaluates all currently blocked actions to determine if any of the newly observed objects could help resolve any blockage. 

% Since all movable objects of a certain size possess the property that moving them may reveal objects of a smaller size behind them, actions such as \texttt{placeItemElsewhereForVisibility(x, y)} and \texttt{nudgeItemForVisibility(x, y)} are added to the tree structure for every item \texttt{x} in the fridge that could possibly cover the milk and all potential destinations \texttt{y} as possible resolutions for the milk not being visible. 

The selection engine opts for the \texttt{\allowbreak{}visualSearch(direction\_bias)} action, this time applying a \texttt{\allowbreak{}direction\_bias} toward the refrigerator. The robot maneuvers its viewpoint and locates the milk in the back of the refrigerator. The blocking condition regarding the milk's visibility is thereby resolved, and all other resolution actions related to this issue are consequently removed from the tree structure.

The action \texttt{\allowbreak{}grasp(milk)} is no longer obstructed, and the selection engine elects to pursue this action next.  After the \texttt{grasp} action is executed, the blocking condition affecting the \texttt{place} action has been cleared. Consequently, the robot positions the milk on the counter, successfully accomplishing its given goal.

% The example above highlights how our proposed strategy nicely models this action selection problem.  For instance, the robot here is able to make one-at-a-time decisions in real-time, quickly adapting to changing physical or information blockages without getting bogged down by all possible actions that are possibly viable along the way.  

% This adaptable strategy even enables the robot to decide when and how it will get more information from the world, such as when the robot looks around the kitchen and the refrigerator to find the milk.  

\section{Initial Instantiation of Strategy}
\label{sec:technical_details}

In this section, we present an initial instance of the strategy discussed in \S\ref{sec:strategy_overivew}.  We also employ this implementation in our evaluation in \S\ref{sec:evaluation}.

% At a high level, our strategy interleaves adding edges and nodes to a state-transition graph (illustrated in Figure \ref{fig:tree_structure}) and using a selection engine to decide on the next action.  The state-transition graph is implemented using standard graphical programming patterns, and the selection engine uses a Large Language Model (LLM) approach.

% Here, we provide a comprehensive explanation of our algorithm, with a particular focus on the design and organization of the selection engine.

\subsection{Selection Engine using LLM}
\label{sec:selection_engine}
As outlined in Section \ref{sec:strategy_steps}, an integral component of our proposed strategy involves a selection engine tasked with choosing an action from a set of candidate actions.  In this work, we use an LLM to make this selection, aiming to use its understanding of context to effectively advance the robot toward its goal.  

% In this work, the selection engine leverages an LLM.  Our methodology is designed to be compatible with any standard LLM and operates in a zero-shot manner, eliminating the need for additional training or fine-tuning.

% \textcolor{blue}{The state transition graph comprises a set of resolution functions and failure-tracking mechanisms. Each goal literal has an associated resolution function, which verifies the current system state against the required predicates at each stage in the transition graph. For instance, the resolution function for the "onTop" predicate checks whether the object is held, and whether the goal surface is near the robot, and, if any of these conditions are not met, it subsequently invokes the resolution function for each unmet predicate.}

At each step, the algorithm first aggregates all candidate actions from which the selection action can be chosen.  This set of candidate actions is comprised of all leaf nodes in the state-action graph, which represent all possible actions that have effects that may either directly achieve a specified goal or resolve a current blocking condition.  For instance, in Figure \ref{fig:tree_structure}, the set of feasible actions would be \texttt{[visualSearch(direction\_bias), open(oven), open(refrigerator), open(freezer)]}.  All of these actions may resolve the blocking condition ``milk must be seen''.   

% \textcolor{blue}{Given the system's modality of selecting one action at a time, there is an increased risk of the system becoming stuck in cycles. Repeated actions are removed from the proposed action set to mitigate this risk. Additionally, suppose the same failure mode is encountered multiple times in close succession during execution. In that case, the action set provided to the LLM is expanded to include all feasible actions in the current state.}

The set of all candidate actions is converted into a text string.  This string is then incorporated into an expansive prompt, the specifics of which are elaborated upon in Section \ref{sec:prompt_gen}. The fully constructed prompt is subsequently dispatched to the LLM. Following the LLM's analysis of the prompt, it produces a response articulated in natural language. The action chosen by the LLM is identified and extracted from this output.

% To simplify the extraction process, the LLM is instructed to format its reply according to a predefined structure (detailed below).

Steps 2 and 3 in \S\ref{sec:strategy_steps} progress with the action selected by the LLM module.  When the approach circles back to Step 1 (the selection engine step), a new prompt is generated given the current context.  Every decision the LLM makes regarding action selection is conducted afresh without leveraging any knowledge from its prior interactions or choices.  While we speculate that maintaining continuous interaction with the LLM throughout all decisions may significantly impact the observed results, these investigations are beyond the scope of our current work.  We discuss this point further in \S\ref{sec:discussion}.    

% At the heart of our robotic system lies an action selection mechanism, leveraging the capabilities of an integrated LLM to pinpoint the action most conducive to advancing the robot toward its designated goal. The representation of actions within our system is succinct, encapsulated within 2-3 token-long strings. These strings begin with an action token—comprising 1-2 words joined by underscores—followed by one or more tokens specifying any instances (nouns) pertinent to the action. To illustrate, an action intending to place a mug into a sink is denoted as ``put\_in mug sink''.

% The process of determining an action's relevance to the ongoing task and the robot's current state involves a reasoning mechanism over the active predicates within said state. This mechanism, integral to our planner, identifies the predicates for evaluation by referencing the current goal predicate of the robot. Each goal predicate is aligned with a distinct resolution function. These resolution functions, structured hierarchically, invoke subordinate predicate resolution functions essential for the execution of the corresponding action.

% Upon the assessment of these predicates, the resolution functions return a set of task-relevant actions that are feasibly executable within the present state, taking into account the necessary preconditions for each action. These selected actions are then formulated into a prompt, which is subsequently sent to the LLM. The LLM, in turn, renders its decision, selecting an action from the presented set, and maintaining the established action representation format.

\subsection{LLM Prompt Generation}
\label{sec:prompt_gen}
% The LLM prompt generation method in our approach involves populating a predefined template with detailed information at every decision.  Given that the LLM does not maintain a history of prior interactions or decisions, it is imperative to continuously reiterate the LLM's role within our framework and provide sufficient context. This ensures the LLM is consistently reminded of its function, particularly focusing on aiding in selecting actions for a robot operating within a domestic setting. Below, we delineate the eight ordered components integrated into the prompt; a full example prompt can be seen in the Appendix:

In our approach, generating prompts for the LLM involves populating a predefined template with specific details at each decision point.  Given that the LLM does not maintain a history of prior interactions or decisions, it's crucial to repeatedly clarify its role and provide ample context within our framework.  We outline below the eight sequential components that make up the prompt, with complete input and output examples in Appendix \ref{app:llm_prompt} and \ref{app:llm_output}:

(1) The prompt begins by stating the LLM's role, indicating that the LLM is assisting the user in selecting their next action. This sets a clear context and defines the scope of tasks expected to be addressed; (2) The prompt next incorporates instructions for the LLM, advising it to take its time and reason methodically, aligning with recommendations from the literature \cite{kojima2023large}; (3) An array of previous actions taken by the robot is added to the prompt.  This summary gives a sense of history in the interaction to contextualize the current decision; (4) An array of sub-goals already completed by the robot is added to the prompt.  Again, this list provides important context for the current decision; (5) The array of candidate actions is added to the prompt; (6) An array of remaining goal predicates are added to the prompt, focusing the robot's effort towards achieving its given goals; (7) Next, the prompt includes descriptions of any errors or failures encountered during previous actions.  These descriptions may include relevant errors returned by the simulation in action execution, any unforeseen blocking condition encountered, or if the action returned by the LLM was not part of the list of candidate actions provided; and (8) Finally, the prompt again emphasizes the necessity of selecting actions from the provided list and asks the LLM output its response in a specified output format using the sentence \texttt{provide your selected action in the format 'format '\$\$ <selected action> \$\$}.  This format string makes it easier to extract the selected action from the output.

These ordered components were selected based on extensive trial-and-error testing.  While a goal of this work is to show that these prompt components are sufficient for effective action selection, demonstrating the necessity or optimality of these (or any) prompt components is beyond the scope of our current investigation.  

% As we will demonstrate in our evaluation, this structured approach to prompt generation appears to be a promising method for enabling the LLM to make effective one-at-a-time action selection decisions.  

\section{Evaluation}
\label{sec:evaluation}

\begin{comment}
\begin{figure*}[t!]
	\includegraphics[width=\textwidth]{figures/placeholder_page_width.png}
	\caption{}
	\label{wide}
	\vspace{-12pt}
\end{figure*}
\end{comment}

In this section, we present an evaluation of our sequential action selection approach, comparing its performance against various LLM and traditional task-planning alternatives. 

% We discuss the implications of our findings in \S\ref{sec:discussion}.    

%  Below, we provide the specifics of our implementation, detail the experimental framework, overview the baseline comparisons, discuss the metrics used for evaluation, and present the final results.  

% Our evaluation is conducted within the AI2Thor environment, focusing on kitchen scenarios that introduce various blocking conditions, to test the adaptability and efficiency of each approach.

%  under constrained and dynamic conditions

%  comparing its performance against two notable methodologies: ProgPrompt \cite{singh2022progprompt}, a few-shot Large Language Model (LLM) planner, and a modified version of the FF-replan \cite{yoon2007ff}

\subsection{Implementation Details}
Our experimental implementation is programmed in Python.  Experiments were run on an Asus Vivobook laptop with a 2.4 GHz Intel Core i7 processor and 16GB RAM.  While our selection engine approach posed in \S\ref{sec:technical_details} can with any off-the-shelf LLM, our current implementation is integrated with OpenAI LLMs.  Because our approach needs to make fast, real-time decisions, our evaluation uses the efficient GPT-3.5 Turbo model \cite{achiam2023gpt}.      

\subsection{Experimental Testbed}
Our experimental testbed is set within the AI2Thor simulation environment \cite{kolve2017ai2}, designed to simulate realistic home scenarios for AI agents.  

% \footnote{\href{https://ai2thor.allenai.org/}{https://ai2thor.allenai.org/}}

The experimental testbed is structured as follows: 

(1) An experimental condition (i.e., some action selection approach) is provided a goal literal.  The process of trying to achieve a goal will be called a ``task''.  The approach selects a sequence of actions, trying to achieve the provided goal within the simulation environment.  Each approach only interfaces with the simulator through action selection, all other rendering and object manipulation are handled by the simulator.  We record information and metrics along the way.  If the goal is not reached within 100 actions, this trial is considered a failure for the given approach; (2) Step 1 is repeated for 50 trials, with information and metrics recorded for all trials; (3) Steps 1 -- 2 are repeated for four tasks, each characterized by their own goal literal (outlined below); and (4) Steps 1 -- 3 are repeated for five action selection conditions (outlined below). 

% multiple potential blocking conditions

\subsection{Baseline Comparisons}
\label{sec:baseline_comparisons}
We compare our approach (Blocking-conditions and Resolutions Action Selection, abbreviated as BCR) to two baselines: (1) ProgPrompt, a few-shot Large Language Model (LLM) planner \cite{singh2022progprompt}; and (2) FF-Replan \cite{yoon2007ff}.  We chose these comparisons because they represent two common yet disparate strategies for discrete action selection.   

For both ProgPrompt and FF-Replan, we consider two conditions each.  We use a publicly available implementation of ProgPrompt.\footnote{\href{https://github.com/NVlabs/progprompt-vh}{https://github.com/NVlabs/progprompt-vh}}  The algorithm presented by \citet{singh2022progprompt} includes an action \texttt{find ?obj}, where \texttt{?obj} is some stand-in variable for an object in the environment.  As the name implies, this action results in the robot automatically finding a certain object in the environment.  The two conditions for ProgPrompt are (1) the exact version proposed by \cite{singh2022progprompt} (referred to as \textit{ProgPrompt}; and (2) a version that excludes the \texttt{find ?obj} action (referred to as \textit{ProgPrompt-no-find}).  This distinction allows us to assess ProgPrompt's performance under optimal conditions as well as in scenarios, akin to our approach, where this automatic locating of objects is absent and objects must be located manually using actions like \texttt{rotate\_right} or \texttt{rotate\_left}.    

We also use a publicly available implementation of FF-Replan.\footnote{\href{https://fai.cs.uni-saarland.de/hoffmann/cff.html}{https://fai.cs.uni-saarland.de/hoffmann/cff.html}}  The conditions are (1) the standard version presented by \citet{yoon2007ff} (referred to as \textit{FF-Replan}), and (2) a version that removes all room location information for relevant goal objects before attempting to generate plans (referred to as \textit{FF-Replan-limited}).  Similar to the case above, this distinction allows us to assess FF-Replan's performance under optimal conditions as well as in scenarios, akin to our approach, where the planner has to locate pertinent objects on the fly.

% In addition to FF-Replan, and ProgPrompt we also include evaluation in comparison of ablations of these algorithms. In the case of FF-Replan, we compare our system against FF-Replan-limited for which we remove all room location information for relevant goal objects before attempting to generate plans. For ProgPrompt we additionally compare our system against ProgPrompt-NF which limits the benefits of the "find ?obj" function utilized by ProgPrompt in that it visually scans the area to see if the target object is visible from the robot's current position, but does not navigate to the object regardless of where it is in the scene. 

\subsection{Experimental Tasks}

Each action selection condition (outlined below) was tasked with executing four distinct tasks within the experimental testbed:

\noindent 1) \textit{Making Coffee} (abbreviated as \textit{Coffee}): involves locating a mug and a coffee maker and turning the coffee maker on while the mug is in the output area.

\noindent 2) \textit{Putting the Apple in the Fridge} (abbreviated as \textit{Apple}): Involves the robot locating the apple and fridge, and storing an apple in the fridge.

\noindent 3) \textit{Washing the Mug in the Sink} (abbreviated as \textit{Mug}): Involves the robot locating the mug and sink, putting the mug in the sink, and turning the faucet on.

\noindent 4)  \textit{Making Toast} (abbreviated as \textit{Toast}): involves locating a loaf of bread and a toaster, slicing the bread, and toasting it in the toaster.

The tasks above are specified by goal literals, and provided to each condition at the beginning of each trial.  

For all tasks, necessary components are factored ahead of time to be compatible with each condition in \S\ref{sec:baseline_comparisons}.  For instance, actions are manually associated with blocking conditions and effects for our approach, actions are manually associated with preconditions and effects for FF-Replan, and literals are reformatted into Python function templates for ProgPrompt.  All possible actions are also associated with their own AI2Thor wrapper function.  AI2Thor then knows how to execute each of these wrapper functions at run-time to correctly update the simulation environment.  

Importantly, certain actions within these tasks are subject to points of failure.  For example, the ``navigate to object'' action does not take into account the optimal point of interaction with the object, so if the robot navigates to the fridge and then opens it, the fridge door may block the robot's view not allowing it to interact with objects in the fridge.  Thus, a key aspect of our evaluation is assessing the resilience of various action selection approaches in the face of such obstacles or uncertainties.  

% The actions provided to the planners are represented in an AI2Thor wrapper as functions. These functions emulate low-level planning systems guiding the robot's execution. The functions are all subject to points of failure, for example, the "navigate to object" function does not take into account the optimal point of interaction with the object so if the robot navigates to the fridge, and then opens the fridge the fridge door may block the robot's view not allowing it to interact with objects in the fridge. 

\begin{table}[t]
    \centering
    % \caption{Success Rate}
    \caption{\textbf{Table 1}: Success Rate}
    \begin{tabular}{c|c|c|c|c}
         &  \textbf{Coffee} & \textbf{Apple} & \textbf{Mug} & \textbf{Toast} \\
         \hline 
        {\scriptsize \textit{BCR (ours)}} & 50/50 & 39/50 & 39/50 & 49/50 \\
        \hline
        {\scriptsize \textit{ProgPrompt}} & 19/50 & 40/50 & 18/50 & 0/50 \\
        \hline 
        {\scriptsize \textit{FF-Replan}} & 14/50 & 13/50 & 11/50 & 0/50 \\
        \hline 
        {\scriptsize \textit{ProgPrompt-no-find}} & 14/50 & 12/50 & 0/50 & 0/50 \\
        \hline
        {\scriptsize \textit{FF-Replan-limited}} & 0/50 & 0/50 & 0/50 & 0/50 \\
        \hline
    \end{tabular}
    \label{tab:table1}
    % \vspace{-15pt}
\end{table}

\begin{table}[t]
    \centering
    % \caption{Success Rate}
    \caption{\textbf{Table 2}: Mean Runtime per Trial}
    \begin{tabular}{c|c|c|c|c}
         &  \textbf{Coffee} & \textbf{Apple} & \textbf{Mug} & \textbf{Toast} \\
         \hline 
        {\scriptsize \textit{BCR (ours)}} & {\scriptsize $10.6 {\tiny \pm} 16$} & {\scriptsize $46.8 {\tiny \pm} 50$} & {\scriptsize $46.4 {\tiny \pm} 44$} & {\scriptsize $11.6 {\tiny \pm} 16$} \\
        \hline
        {\scriptsize \textit{ProgPrompt}} & {\scriptsize $20.3 {\tiny \pm} 6.7$} & {\scriptsize $8.77 {\tiny \pm} 8.4$} & {\scriptsize $73.8 {\tiny \pm} 75$} & {\scriptsize $24.1 {\tiny \pm} 3.1$} \\
        \hline 
        {\scriptsize \textit{FF-Replan}} & {\scriptsize $0.52 {\tiny \pm} 0.3$} & {\scriptsize $0.97 {\tiny \pm} 0.4$} & {\scriptsize $6.90 {\tiny \pm} 5$} & {\scriptsize $2.74 {\tiny \pm} 5.6$} \\
        \hline 
        {\scriptsize \textit{ProgPrompt-no-find}} & {\scriptsize $20.2 {\tiny \pm} 5.5$} & {\scriptsize $20.7 {\tiny \pm} 9.2$} & {\scriptsize $117 {\tiny \pm} 59$} & {\scriptsize $27.9 {\tiny \pm} 41$} \\
        \hline
        {\scriptsize \textit{FF-Replan-limited}} & {\scriptsize $4.23 {\tiny \pm} 0.55$} & {\scriptsize $7.81 {\tiny \pm} 1.7$} & {\scriptsize $8.8 {\tiny \pm} 2.6$} & {\scriptsize $5.4 {\tiny \pm} 1.8$} \\
        \hline
    \end{tabular}
    \caption{\small (\textit{measured in seconds and range values are standard deviation})}
    \label{tab:table2}
    \vspace{-20pt}
\end{table}

\begin{table}[t]
    \centering
    % \caption{Success Rate}
    \caption{\textbf{Table 3}: Mean Number of Actions Considered per Decision}
    \begin{tabular}{c|c|c|c|c}
         &  \textbf{Coffee} & \textbf{Apple} & \textbf{Mug} & \textbf{Toast} \\
         \hline 
        {\scriptsize \textit{BCR (ours)}} & {\scriptsize $11.2 {\tiny \pm} 7.6$} & {\scriptsize $12.0 {\tiny \pm} 4.5$} & {\scriptsize $13.6 {\tiny \pm} 4.5$} & {\scriptsize $11.0 {\tiny \pm} 5.4$} \\
        \hline
        {\scriptsize \textit{FF-Replan}} & {\scriptsize $49 {\tiny \pm} 56$} & {\scriptsize $117 {\tiny \pm} 118$} & {\scriptsize $139 {\tiny \pm} 135$} & {\scriptsize $135 {\tiny \pm} 117$} \\
        \hline 
        {\scriptsize \textit{FF-Replan-lim.}} & {\scriptsize $661 {\tiny \pm} 74$} & {\scriptsize $649 {\tiny \pm} 112$} & {\scriptsize $721 {\tiny \pm} 213$} & {\scriptsize $669 {\tiny \pm} 106$} \\
        \hline
    \end{tabular}
    \caption{\small (\textit{range values are standard deviation})}
    \label{tab:table3}
    \vspace{-20pt}
\end{table}

\subsection{Evaluation Metrics}

We report on three metrics in our evaluation: (1) success rate, the number of trials out of 50 total trials where the approach was able to reach the goal; (2) The average runtime per trial; and (3) the average number of actions considered per action selection decision (not applicable for ProgPrompt).  

% Note that the third metric does not apply for the ProgPrompt conditions since this purely LLM-based planner does not consider a certain number of actions per decision.

\subsection{Results}
As shown in Table 1, our approach has a higher success rate than the alternative conditions.  Even when ProgPrompt and FF-Replan are provided additional information or enhanced actions not available to our approach, we see that our approach still has a comparable or higher success rate.  Moreover, as demonstrated in Table 2, the runtime of our approach is comparable to ProgPrompt, although it is not as fast as FF-Replan.  Each decision in our approach takes about one second, which is certainly fast enough to make one-at-a-time sequential decisions in real time. 

Additionally, Table 3 shows that our approach achieves these results while considering many fewer actions than the FF-Replan conditions.  This suggests that our approach is more data efficient, helping to mitigate the exponential scaling of action branches typically exhibited by task planners.   

% As shown in Table I, our method outperforms both standard ProgPrompt in all but one setting, and our modified FF-Replan in all experimental settings, these results are emphasised in the limited implementations of each of these algorithms. Our system proved to be more adaptable to blocking scenarios and was able to work around blocking conditions where the other methods were not.
\section{Discussion}
\label{sec:discussion}

In this work, we have presented a strategy that frames the action selection problem for robots in terms of resolving blocking conditions.  This strategy allows a robot to make one-at-a-time decisions that take context and adapt and react to current situations.  We also presented the first implementation of this strategy that uses a natural language processing model (LLM) as the selection engine. Our tests in a simulated environment show that our approach is often more effective than alternative methods, leading to a higher success rate in completing various tasks.

\subsection{Limitations and Future Work}

We note several limitations of our work that suggest future extensions and investigations.  First, the LLM selection engine approach in \S\ref{sec:technical_details} and prompt components in \S\ref{sec:prompt_gen} reflect just one possible instance of the strategy specified in \S\ref{sec:strategy_overivew}.  Our current work does not claim that any of these choices are ``best'', only sufficient within the context of our evaluation.  We will continue to explore this broad design space going forward, assessing the necessity and optimality of these components.

% of natural language solutions within a blocking conditions and resolutions-based action selection approach, as well as proposing and comparing alternative prompt structures.

The results in \S\ref{sec:evaluation} suggest that our approach scales well to tens of candidate actions (i.e., $<100$) at a time.  However, the LLM would likely get overwhelmed with decisions on the order of hundreds or thousands of candidate actions.  Additional investigation is needed to characterize and address these possible scaling challenges.
% were selected based on extensive testing; however, they

% \textcolor{blue}{A limitation of our evaluation is the exclusive comparison to ProgPrompt as an LLM planner. We did not include other state-of-the-art planners due to their reliance on more detailed environmental information, such as the affordance function used by the SayCan approach \cite{ahn2022i}. Future iterations should address this by incorporating comparisons with SayCan and other advanced methods, including chain-of-thought type planners \cite{wei2022chainofthought}. }

Our approach outlined in \S\ref{sec:technical_details} exhibits several occasional errors.  For instance, even when directly instructed to only select an action from the set of candidate actions, the LLM may still hallucinate and select an action from outside of this set.  Our current implementation detects this error and simply re-sends the prompt with an additional note urging the LLM to carefully read this instruction, which often fixes this issue.  Also, it is possible for the LLM to get caught in a loop of repeated actions.  Many of the unsuccessful trials in our evaluation were due to this issue, wherein our approach reached the 100-action maximum.  We speculate that this issue may be mitigated by maintaining continuous interaction with the LLM instead of starting a fresh interaction at every decision.  We plan to directly address this challenge in future work.

Our approach currently requires all actions to be manually populated with blocking conditions and resolutions.  While this process is analogous to manually specifying preconditions and effects for standard task planners, the process may still be tedious.  We plan to investigate methods to automatically generate these connections between actions, blocking conditions, and resolutions, both with offline pre-processing as well as creative online inference. A promising future direction is to incorporate Visual Language Models (VLMs) capable of detecting and adapting to previously unknown blocking modes encountered during runtime.

\subsection{LLMs for Sequential Task Reasoning}

Our work demonstrates a promising application of LLMs.  We show that, if used in a particular way, even LLM models that are not the largest or newest at the time of writing (e.g., GPT-3.5) can enable effective sequential discrete action selection.  This observation is in contrast to other works that suggest that even more modern LLM models (e.g., GPT-4) still struggle with sequential task reasoning \cite{bubeck2023sparks}. 

We suggest the following hypothesis: \textit{current LLM models may struggle with robustly formulating a sequence of actions to reach a goal all in one output.  However, these models may excel at making one-at-a-time action selection decisions to reach a specified goal if given choices and enough context at each decision point}.  While our current work does not provide enough evidence to prove this point, our findings offer preliminary evidence suggesting some validity of this phenomenon.  We suggest this hypothesis as an exciting avenue of research going forward.  

% Going forward, we plan to incorporate our approach with continuous motion generation, affording robust action selection between non-linear optimization loops.  

% \lipsum[5-10]

%%%%%%%%%%%%%%%%%%%%%%%%%%%%%%%%%%%%%%%%%%%%%%%%%%%%%%%%%%%%%%%%%%%%%%%%

%%%%%%%%%%%%%%%%%%%%%%%%%%%%%%%%%%%%%%%%%%%%%%%%%%%%%%%%%%%%%%%%%%%%%%%

\section*{ACKNOWLEDGMENT}

The authors thank Rachel Holladay for interesting and insightful conversations regarding this work.

%%%%%%%%%%%%%%%%%%%%%%%%%%%%%%%%%%%%%%%%%%%%%%%%%%%%%%%%%%%%%%%%%%%%%%%

\bibliographystyle{plainnat}

\bibliography{refs}

\begin{thebibliography}{19}
\providecommand{\natexlab}[1]{#1}
\providecommand{\url}[1]{\texttt{#1}}
\expandafter\ifx\csname urlstyle\endcsname\relax
  \providecommand{\doi}[1]{doi: #1}\else
  \providecommand{\doi}{doi: \begingroup \urlstyle{rm}\Url}\fi

\bibitem[Achiam et~al.(2023)Achiam, Adler, Agarwal, Ahmad, Akkaya, Aleman, Almeida, Altenschmidt, Altman, Anadkat, et~al.]{achiam2023gpt}
Josh Achiam, Steven Adler, Sandhini Agarwal, Lama Ahmad, Ilge Akkaya, Florencia~Leoni Aleman, Diogo Almeida, Janko Altenschmidt, Sam Altman, Shyamal Anadkat, et~al.
\newblock Gpt-4 technical report.
\newblock \emph{arXiv preprint arXiv:2303.08774}, 2023.

\bibitem[Ahn et~al.(2022)Ahn, Brohan, Brown, Chebotar, Cortes, David, Finn, Fu, Gopalakrishnan, Hausman, Herzog, Ho, Hsu, Ibarz, Ichter, Irpan, Jang, Ruano, Jeffrey, Jesmonth, Joshi, Julian, Kalashnikov, Kuang, Lee, Levine, Lu, Luu, Parada, Pastor, Quiambao, Rao, Rettinghouse, Reyes, Sermanet, Sievers, Tan, Toshev, Vanhoucke, Xia, Xiao, Xu, Xu, Yan, and Zeng]{ahn2022i}
Michael Ahn, Anthony Brohan, Noah Brown, Yevgen Chebotar, Omar Cortes, Byron David, Chelsea Finn, Chuyuan Fu, Keerthana Gopalakrishnan, Karol Hausman, Alex Herzog, Daniel Ho, Jasmine Hsu, Julian Ibarz, Brian Ichter, Alex Irpan, Eric Jang, Rosario~Jauregui Ruano, Kyle Jeffrey, Sally Jesmonth, Nikhil~J Joshi, Ryan Julian, Dmitry Kalashnikov, Yuheng Kuang, Kuang-Huei Lee, Sergey Levine, Yao Lu, Linda Luu, Carolina Parada, Peter Pastor, Jornell Quiambao, Kanishka Rao, Jarek Rettinghouse, Diego Reyes, Pierre Sermanet, Nicolas Sievers, Clayton Tan, Alexander Toshev, Vincent Vanhoucke, Fei Xia, Ted Xiao, Peng Xu, Sichun Xu, Mengyuan Yan, and Andy Zeng.
\newblock {Do As I Can, Not As I Say: Grounding Language in Robotic Affordances}, 2022.

\bibitem[Bertoli et~al.(2002)Bertoli, Cimatti, et~al.]{bertoli2002improving}
Piergiorgio Bertoli, Alessandro Cimatti, et~al.
\newblock Improving heuristics for planning as search in belief space.
\newblock In \emph{AIPS}, pages 143--152, 2002.

\bibitem[Bonet and Geffner(2000)]{bonet2000planning}
Blai Bonet and Hector Geffner.
\newblock Planning with incomplete information as heuristic search in belief space.
\newblock In \emph{Proceedings of the Fifth International Conference on Artificial Intelligence Planning Systems}, pages 52--61, 2000.

\bibitem[Bubeck et~al.(2023)Bubeck, Chandrasekaran, Eldan, Gehrke, Horvitz, Kamar, Lee, Lee, Li, Lundberg, et~al.]{bubeck2023sparks}
S{\'e}bastien Bubeck, Varun Chandrasekaran, Ronen Eldan, Johannes Gehrke, Eric Horvitz, Ece Kamar, Peter Lee, Yin~Tat Lee, Yuanzhi Li, Scott Lundberg, et~al.
\newblock Sparks of artificial general intelligence: Early experiments with gpt-4.
\newblock \emph{arXiv preprint arXiv:2303.12712}, 2023.

\bibitem[Chalvatzaki et~al.(2023)Chalvatzaki, Younes, Nandha, Le, Ribeiro, and Gurevych]{chalvatzaki2023learning}
Georgia Chalvatzaki, Ali Younes, Daljeet Nandha, An~Le, Leonardo F.~R. Ribeiro, and Iryna Gurevych.
\newblock Learning to reason over scene graphs: A case study of finetuning gpt-2 into a robot language model for grounded task planning, 2023.

\bibitem[Ding et~al.(2022)Ding, Zhang, Amiri, Cao, Yang, Esselink, and Zhang]{ding2022robot}
Yan Ding, Xiaohan Zhang, Saeid Amiri, Nieqing Cao, Hao Yang, Chad Esselink, and Shiqi Zhang.
\newblock Robot task planning and situation handling in open worlds, 2022.

\bibitem[Fikes and Nilsson(1971)]{fikes1971strips}
Richard~E Fikes and Nils~J Nilsson.
\newblock Strips: A new approach to the application of theorem proving to problem solving.
\newblock \emph{Artificial intelligence}, 2\penalty0 (3-4):\penalty0 189--208, 1971.

\bibitem[Helmert(2006)]{helmert2006fast}
Malte Helmert.
\newblock The fast downward planning system.
\newblock In \emph{Journal of Artificial Intelligence Research}, pages 191--246, 2006.

\bibitem[Hoffmann and Brafman(2005)]{hoffmann2005contingent}
J{\"o}rg Hoffmann and Ronen~I Brafman.
\newblock Contingent planning via heuristic forward search with implicit belief states.
\newblock In \emph{Proceedings of the International Conference on Automated Planning and Scheduling}, volume~15, pages 71--80, 2005.

\bibitem[Hoffmann and Nebel(2001)]{hoffmann2001ff}
J{\"o}rg Hoffmann and Bernhard Nebel.
\newblock Ff: The fast-forward planning system.
\newblock \emph{AI Magazine}, 22\penalty0 (3):\penalty0 57, 2001.

\bibitem[Kojima et~al.(2023)Kojima, Gu, Reid, Matsuo, and Iwasawa]{kojima2023large}
Takeshi Kojima, Shixiang~Shane Gu, Machel Reid, Yutaka Matsuo, and Yusuke Iwasawa.
\newblock Large language models are zero-shot reasoners, 2023.

\bibitem[Kolve et~al.(2017)Kolve, Mottaghi, Han, VanderBilt, Weihs, Herrasti, Deitke, Ehsani, Gordon, Zhu, et~al.]{kolve2017ai2}
Eric Kolve, Roozbeh Mottaghi, Winson Han, Eli VanderBilt, Luca Weihs, Alvaro Herrasti, Matt Deitke, Kiana Ehsani, Daniel Gordon, Yuke Zhu, et~al.
\newblock Ai2-thor: An interactive 3d environment for visual ai.
\newblock \emph{arXiv preprint arXiv:1712.05474}, 2017.

\bibitem[Liu et~al.(2023)Liu, Jiang, Zhang, Liu, Zhang, Biswas, and Stone]{liu2023llmp}
Bo~Liu, Yuqian Jiang, Xiaohan Zhang, Qiang Liu, Shiqi Zhang, Joydeep Biswas, and Peter Stone.
\newblock Llm+p: Empowering large language models with optimal planning proficiency, 2023.

\bibitem[McDermott et~al.(1998)McDermott, Ghallab, Howe, Knoblock, Ram, Veloso, Weld, and Wilkins]{mcdermott1998pddl}
Drew McDermott, Malik Ghallab, Adele Howe, Craig Knoblock, Ashwin Ram, Manuela Veloso, Daniel Weld, and David Wilkins.
\newblock Pddl - the planning domain definition language.
\newblock In \emph{AIPS98 Planning Committee}. Yale Center for Computational Vision and Control, 1998.

\bibitem[Singh et~al.(2022)Singh, Blukis, Mousavian, Goyal, Xu, Tremblay, Fox, Thomason, and Garg]{singh2022progprompt}
Ishika Singh, Valts Blukis, Arsalan Mousavian, Ankit Goyal, Danfei Xu, Jonathan Tremblay, Dieter Fox, Jesse Thomason, and Animesh Garg.
\newblock Progprompt: Generating situated robot task plans using large language models, 2022.

\bibitem[Song et~al.(2023)Song, Wu, Washington, Sadler, Chao, and Su]{song2023llmplanner}
Chan~Hee Song, Jiaman Wu, Clayton Washington, Brian~M. Sadler, Wei-Lun Chao, and Yu~Su.
\newblock Llm-planner: Few-shot grounded planning for embodied agents with large language models, 2023.

\bibitem[Yoon et~al.(2007)Yoon, Fern, and Givan]{yoon2007ff}
Sungwook Yoon, Alan Fern, and Robert Givan.
\newblock Ff-replan: A baseline for probabilistic planning.
\newblock \emph{Artificial Intelligence}, 14:\penalty0 255--279, 2007.

\bibitem[Zhao et~al.(2023)Zhao, Zhou, Li, Tang, Wang, Hou, Min, Zhang, Zhang, Dong, et~al.]{zhao2023survey}
Wayne~Xin Zhao, Kun Zhou, Junyi Li, Tianyi Tang, Xiaolei Wang, Yupeng Hou, Yingqian Min, Beichen Zhang, Junjie Zhang, Zican Dong, et~al.
\newblock A survey of large language models.
\newblock \emph{arXiv preprint arXiv:2303.18223}, 2023.

\end{thebibliography}

% \begin{thebibliography}{99}
% \end{thebibliography}

\appendix
\subsection{Example LLM Prompt}
\label{app:llm_prompt}
Here, we provide a full example prompt dispatched to the LLM in our approach.  (Note that tags like \texttt{Cabinet|-01.14|+00.39|+03.52} are object IDs for instances in AI2Thor).

\begin{lstlisting}[escapeinside={(*@}{@*)}]
(*@\seqsplit{[{'role': 'system', 'content': 'You are helping me select my next action, take your time and verify that the action you select is part of the list I provide. Take your time and go step by step.'}, {'role': 'user', 'content': 'I am in the kitchen. I am not holding anything.'}, {'role': 'user', 'content': 'The action scanroom if available allows me to visually scan the room I am currently in to see if an object is visible. All actions require the target object to be visible other than scanroom.'}, {'role': 'user', 'content': 'Some objects may be contained within others that need to be opened and wont be found by scanroom.'}, {'role': 'user', 'content': "select the the best action from this list: ['\$\$ turnleft character \$\$', '\$\$ turnright character \$\$', '\$\$ moveforward \$\$', '\$\$ movebackward \$\$', '\$\$ turnaround \$\$', '\$\$ lookup \$\$', '\$\$ lookdown \$\$', '\$\$ walk\_to\_room kitchen \$\$', '\$\$ open Cabinet|-01.14|+00.39|+03.52 \$\$', '\$\$ walk\_to\_object Drawer|-02.11|+00.71|+03.66 \$\$', '\$\$ walk\_to\_object Cabinet|-02.38|+02.01|+01.69 \$\$', '\$\$ walk\_to\_object Cabinet|-01.98|+01.78|+03.73 \$\$', '\$\$ walk\_to\_object Fridge|-02.64|+00.00|+02.13 \$\$', '\$\$ walk\_to\_object Cabinet|-02.38|+00.64|+01.68 \$\$', '\$\$ walk\_to\_object Cabinet|-02.38|+02.01|+01.06 \$\$', '\$\$ walk\_to\_object Cabinet|-01.96|+01.93|+03.73 \$\$', '\$\$ walk\_to\_object Cabinet|-02.95|+01.93|+03.73 \$\$', '\$\$ walk\_to\_object Cabinet|-02.43|+00.44|+03.42 \$\$', '\$\$ walk\_to\_object Cabinet|-02.38|+01.69|+02.57 \$\$', '\$\$ walk\_to\_object Cabinet|-02.43|+00.44|+03.17 \$\$', '\$\$ walk\_to\_object Cabinet|-02.38|+02.01|+02.56 \$\$', '\$\$ walk\_to\_object Cabinet|-01.99|+00.39|+03.51 \$\$', '\$\$ walk\_to\_object Cabinet|-02.38|+02.01|+03.18 \$\$', '\$\$ walk\_to\_object Cabinet|-02.38|+01.93|+01.69 \$\$', '\$\$ walk\_to\_object Cabinet|-02.38|+01.41|+01.68 \$\$', '\$\$ walk\_to\_object Cabinet|-02.62|+01.93|+03.73 \$\$', '\$\$ walk\_to\_object Cabinet|-01.97|+00.39|+03.52 \$\$', '\$\$ scanroom Apple|-00.74|+00.77|+00.99 kitchen \$\$', '\$\$ turnleft character \$\$', '\$\$ turnright character \$\$', '\$\$ moveforward \$\$', '\$\$ movebackward \$\$', '\$\$ turnaround \$\$', '\$\$ lookup \$\$', '\$\$ lookdown \$\$'] that is most likely to help me achieve my goal: ['INSIDE Apple|-00.74|+00.77|+00.99 Fridge|-02.64|+00.00|+02.13']? include an explaination for your action selection, only use the object name in your consideration ie '<objectname>|\#|\#|\#'. Please refrain from getting stuck in action loops and provide your selected action in the format 'format '\$\$ <selected action> \$\$."}, {'role': 'user', 'content': 'Please only select actions in the list I provided.'}]}@*)
\end{lstlisting}

\subsection{Example LLM Output}
\label{app:llm_output}

Here, we provide a full example output returned by the LLM in our approach:

\begin{lstlisting}[escapeinside={(*@}{@*)}]
(*@\seqsplit{\$\$ walk\_to\_object Fridge|-02.64|+00.00
|+02.13 \$\$.  This action will help you reach the Fridge, which is where the Apple is located based on the information you provided.}@*)
\end{lstlisting}

\subsection{Example Action Sequence}

Here, we provide a full sequence of actions generated by our approach for the task ``putting the apple in the fridge'':

\begin{lstlisting}[escapeinside={(*@}{@*)}]
(*@\seqsplit{['walk\_to\_object Fridge|-02.64|+00.00|+02.13', 'open Fridge|-02.64|+00.00|+02.13', 'scanroom Apple|-00.74|+00.77|+00.99 kitchen', 'grab Apple|-00.74|+00.77|+00.99', 'putin Apple|-00.74|+00.77|+00.99 Fridge|-02.64|+00.00|+02.13', 'put Apple|-00.74|+00.77|+00.99 DiningTable|-00.92|00.00|+01.20', 'open Fridge|-02.64|+00.00|+02.13', 'walk\_to\_object Cabinet|-02.38|+02.01|+01.06', 'open Cabinet|-02.38|+02.01|+01.06', 'scanroom Egg|-02.53|+01.18|+02.22 kitchen', 'grab Egg|-02.53|+01.18|+02.22', 'put Egg|-02.53|+01.18|+02.22 Fridge|-02.64|+00.00|+02.13', 'walk\_to\_room kitchen', 'open Cabinet|-02.38|+01.69|+02.57', 'scanroom Apple|-00.74|+00.77|+00.99 kitchen', 'grab Apple|-00.74|+00.77|+00.99', 'walk\_to\_object Fridge|-02.64|+00.00|+02.13', 'walk\_to\_object Cabinet|+00.58|+00.39|+02.36', 'walk\_to\_room kitchen', 'open Cabinet|+00.58|+00.39|+02.36', 'moveforward', 'movebackward', 'open Drawer|+00.73|+00.70|+02.20', 'scanroom Fridge|-02.64|+00.00|+02.13 kitchen', 'walk\_to\_object Fridge|-02.64|+00.00|+02.13', 'putin Apple|-00.74|+00.77|+00.99 Fridge|-02.64|+00.00|+02.13']}@*)
\end{lstlisting}

\textit{(Note that actions like \texttt{grabEgg} and \texttt{placeEgg} here are resolution actions needed because these objects are in the way, hindering progress on the task at hand)}

\end{document}